\documentclass[12pt,preprint]{elsarticle}

\usepackage[T1]{fontenc}
\usepackage{amsmath,amssymb}
\usepackage{graphicx}
\usepackage{booktabs}
\usepackage{siunitx}
\usepackage{setspace}
\usepackage{algorithm}
\usepackage{textcomp}
\usepackage[english]{babel}
\usepackage[dvipsnames, svgnames, table]{xcolor}
\usepackage{bm}

\usepackage{subcaption}
\usepackage[inline]{enumitem}
\setlist{nolistsep}
\usepackage{hyperref}
\hypersetup{
    breaklinks = true,
    colorlinks = true,
    bookmarksopen = true,
}
\usepackage[section, above, below]{placeins}
\usepackage[noabbrev, nameinlink]{cleveref}

\def\ie{{\textit{i.e.}, }}

\def\be{\begin{equation}}
\def\ee{\end{equation}}
\def\ba{\begin{array}}
\def\ea{\end{array}}

\title{Yarn tracking of large-scale 3D textile reinforcements using topological material features}

\author[1,2,3]{Hafsa El Herichi \corref{cor1}}
\author[1,3]{Arturo Mendoza}
\author[4]{Yanneck Wielhorski}
\author[2]{Hugues Talbot}
\author[1]{Stéphane Roux}
\cortext[cor1]{Corresponding author.}
\ead{hafsa.el_herichi@ens-paris-saclay.fr}

\address[1]{Universit\'e Paris-Saclay, CentraleSup\'elec, ENS Paris-Saclay, CNRS, LMPS, 4, Avenue des Sciences, 91192 Gif-sur-Yvette, France}
\address[2]{Universit\'e Paris-Saclay, CentraleSup\'elec, Inria, CVN, 9 Rue Joliot Curie, 91190 Gif-sur-Yvette, France}
\address[3]{Safran Tech, Rue des Jeunes Bois, 78772 Magny-les-Hameaux, France}
\address[4]{Safran Aircraft Engines, Rond-point R\'en\'e Ravaud - R\'eau, 77550 Moissy-Cramayel, France}

\begin{document}

\begin{abstract}
    Automated segmentation of CT images has become increasingly important to enhance the reliability of simulations through the generation of high fidelity numerical models. This study addresses the challenging task of semi-automatically tracking textile reinforcements in fan blade dry preforms using X-ray CT images captured at coarse resolutions (\ie above 140~{\textmu}m). Our approach offers a scalable, slice-based analysis conducted on planes orthogonal to the main yarn directions, applied to a large-scale real industrial component. This enables accurate identification and tracking of yarn paths while requiring minimal training.

    The method models three key yarn properties statistically: their typical cross-section shape, their continuity and movement in the 3D space, and their spatial relative arrangement with respect to neighboring yarns. These statistical properties are integrated into a tracking framework via
    a variational formulation that optimizes all yarn center positions in successive cross-section planes.

    The method tracks more than 3,000 warp yarns across 1,500 slices and achieves a tracking
    success rate above 90\%. Overall, this work demonstrates a promising approach toward large-scale, automated textile reinforcement annotation, paving the way for more efficient material characterization in complex composite structures.
\end{abstract}

\maketitle


\section{Introduction}

Composite materials have revolutionized modern engineering by offering superior specific mechanical properties compared to traditional materials~\cite{LIU2023111176,Khan_engineering}.

The integration of carbon fiber-reinforced polymers (CFRPs) in aircraft engines has significantly enhanced performance and efficiency~\cite{ZHANG2023110463}.

The introduction of 3D woven composite fan blades and fan case in the LEAP engine contributed to a 15\% reduction in fuel consumption compared to the best-performing engine of the previous generation~\cite{JING2025113262}.

Hence, non-destructive analyses based on X-ray computed tomography~\cite{NDT_sota} have become increasingly important to inspect the internal yarn trajectories, orientations, and spatial arrangements.

Faithful extraction of textile mesoscale geometry is essential for accurate finite element analysis of complex woven architectures~\cite{ZHANG2024110828}. Many studies rely on classical image processing combined with quantitative analyses~\cite{Wielhorski_CompPartA_2022}, including structure tensor methods for yarn orientation estimation~\cite{Naouar_JMS_2020,Fourrier_CompStruct_2023}.
Other approaches include clustering based on grayscale intensity~\cite{Straumit_CompPartA_2015}, texture-based methods using local descriptors~\cite{Naouar_CS_2015}, and variational techniques where an initial 3D yarn geometry is iteratively refined via deformable models to better match the microstructure~\cite{Benezech_CS_2019, Sinchuk_Materials_2020, Brion_CS_2022}.

In recent years, research on textile composite segmentation has largely shifted toward deep learning approaches. These include combining neural networks with watershed segmentation trained on synthetic images to avoid manual labeling~\cite{Sinchuk_CS_2021}. Synthetic data have also been used to train keypoint-based contour detectors on pseudo-CT images derived from finite element yarn labels for direct mesh generation~\cite{Mendoza_CST_2021}, as well as fully convolutional networks applied to simulated CT scans~\cite{FRIEMANN2025112656}.

For real CT datasets, panoptic segmentation with IoU-based tracking enables yarn identification and 3D reconstruction~\cite{Allred2022}, while transformer architectures with multi-scale feature fusion and boundary-guided learning achieve high accuracy in densely packed yarns~\cite{ma18061215}. U-Net-based methods allow accurate segmentation of yarns, matrix, and defects in low-contrast CT~\cite{Chen2021}, and two-stage coarse-to-fine convolutional approaches address heterogeneous woven structures~\cite{Zheng_CS_2024,Zheng_CST_2025}.
Pipelines combining semantic segmentation with object detection identify yarn cross-sections and reconstruct 3D geometries~\cite{Tang_CST_2024}, while multi-stage schemes integrating segmentation, gradient-flow instance separation, and clustering-based tracking reconstruct 3D yarn envelopes from low-contrast CT~\cite{CAO2025113294}.

Other deep-learning approaches~\cite{BLUSSEAU2022110333} leverage manually annotated yarn trajectories and morphology-based pseudo-labels to directly extract yarn centerlines from CT data, rather than segmenting the full structure. This concept is later~\cite{Sinchuk_CompPartA_2024} extended to larger, noisier volumes by converting CT scans into scalar distance maps and vector heat-flow fields, followed by tracking algorithms to reconstruct yarn centerlines.

Classical techniques can be effective when applied to high-resolution, high-contrast images with simple patterns. However, their performance degrades with noise, complex fiber architectures, or low-resolution data. They also require careful parameter tuning and may struggle to capture the continuous, connected nature of yarn structures, limiting scalability to industrial-scale parts.
Meanwhile, deep learning methods have gained popularity due to their adaptability, strong performance in complex pattern recognition, and robustness to noise and low-contrast data. However, despite recent advances in self-supervised learning, they typically require extensive annotated datasets, significant computational resources, and remain limited to specific image classes.

In this article, we propose a novel methodology that directly incorporates prior knowledge of the material structure into the processing pipeline. Rather than relying on implicitly learned descriptors to capture characteristic length scales (e.g., yarn dimensions or spatial arrangements) from limited annotated data, these parameters are explicitly defined based on the known physical architecture and topological constraints of the composite.
By embedding such material-driven features, the approach improves controllability and robustness of the tracking process, ensuring faithful reconstruction of the composite architecture. Moreover, the explicit integration of prior knowledge significantly reduces the amount of annotated data required, which is a major advantage for large-scale CT datasets. The extracted yarn trajectories can then be used both for quality control of textile preforms and as input for mesoscale numerical models of textile reinforcements.

The literature can be grouped into three main resolution ranges. High-resolution studies between 3 and 12~{\textmu}m \cite{Sinchuk_CS_2021, Chen2021, Zheng_CS_2024, CAO2025113294}, typically focused on small-scale material coupons involving only tens of yarns, are followed by intermediate resolutions between 15 and 25~{\textmu}m \cite{Mendoza_CST_2021, Allred2022, ma18061215,Zheng_CST_2025, Tang_CST_2024, BLUSSEAU2022110333}, applied to moderately sized parts, and finally lower resolutions between 40 and 50~{\textmu}m \cite{FRIEMANN2025112656,Sinchuk_CompPartA_2024}, used for larger components.

In addition, these state-of-the-art approaches are evaluated on consolidated composite materials, whereas the present study considers a dry textile preform. Extending the proposed method to injected composite parts is an interesting perspective for future work, requiring dedicated investigation due to different imaging conditions.

While state-of-the-art methods are well-suited for small components, they do not fully capture the true complexity and size of actual industrial parts.
Our work bridges this gap by demonstrating a reliable tracking method applied to CT scans of an actual aeronautical component measuring several tens of centimeters in size and containing over 3,000 individual yarns that must be identified and tracked throughout the volume. The scans were acquired under realistic, challenging imaging conditions resulting in a resolution of 140~{\textmu}m, where typical yarn cross-sections can be as small as $7 \times 10$ voxels with no inner texture (as shown below in \Cref{fig:annotations_full}).

The remainder of the paper first introduces the annotated and semi-automatically generated datasets in \Cref{sec:Material_Dataset}, followed by the modeling of the material structure in \Cref{sec:characterization}. The tracking framework and variational formulation are then presented in \Cref{sec:strategy}, along with the evaluation metrics in \Cref{sec:metrics} and results in \Cref{sec:results}.
Finally, conclusions and perspectives are discussed in \Cref{sec:conclusion}.

\section{Material and dataset}
\label{sec:Material_Dataset}

\subsection{X-ray CT acquisition}

The textile preform was imaged using an industrial X-ray CT system (Phoenix V\textbar tome\textbar x L450) at 300~kV and 600~\textmu A. A total of 2400 projections were acquired over ~30 minutes. The setup (FDD 1300~mm, FOD 910~mm) yielded a magnification of 1.4. Images were captured with a DXR-250 detector ($2024 \times 2024$ pixels, 0.2~mm pitch) and reconstructed using Phoenix datos\textbar x, giving a voxel size of 140~\textmu m. The specimen was scanned in four sections; this study focuses on the blade root region.

\subsection{Manual annotations}

Our analysis focuses on the first scan of the blade root, a structurally critical region connecting the blade to the hub and supporting significant loads. It also features complex weaving, with curved and densely packed yarns, making it a challenging and representative test case.
The scanned volume measures ${796 \times 1848 \times 2383}$ voxels (${111 \times 259 \times 335}$ mm). In the following, only warp yarns are considered. The scan is partially manually annotated.

These annotations were produced manually by expert annotators requiring a substantial amount of time and effort to ensure high-quality ground truth data at this scale. In total, 1170 warp yarns were manually annotated out of an estimated 3057 warp yarns present in the entire volume (\Cref{fig:annotations_full}), representing approximately 38\% of the total.
Each annotated yarn was labeled using a sparse set of discrete points, with a level of discretization chosen to be sufficient to capture the local undulation while minimizing annotation effort. It is important to note that not all yarns are annotated in each 2D slice. Each slice contains only a subset of annotated yarns, reflecting a sparse annotation strategy in both the in-plane and through-thickness directions. Overall, only 30 slices out of 1500 were annotated, corresponding to only 2\% of the available image planes.

Each yarn is represented by a continuous parametric curve describing the evolution of its centerline through the tomographic volume. We introduce a coordinate system $(u,v,w)$, where $\mathbf{e}_w$ denotes the main yarn orientation, and $(\mathbf{e}_u,\mathbf{e}_v)$ span the local cross-sectional plane. For each yarn $i$, the path is defined by $\mathbf{X}_i : \mathbb{R} \rightarrow \mathbb{R}^2,\; w \mapsto \mathbf{X}_i(w) = \big(u_i(w),\, v_i(w)\big)$.

Annotations are sparse, available only every 35 voxels along the yarn trajectories from $w=0$ to $w=1000$. However, the yarn paths are smooth and regular. Therefore, any interpolation (e.g., linear or cubic spline) between annotated planes is sufficient to estimate intermediate positions faithfully at non-annotated slices $w$.

The functions $u_i(w)$ and $v_i(w)$ are treated as parametric functions of the slice position $w$, providing a continuous description of each yarn centerline within the cross-sectional plane.
The domain of $\mathbf{X}_i$ corresponds to the set of discrete tomographic slices. The complete yarn system, composed of $N_t$ yarns of the same family (warp or weft), is then expressed as the ordered set $\mathbf{X}(w) = \{\mathbf{X}_i(w)\}_{i=1}^{N_t}$.

This formalism provides a compact and consistent way to describe the evolution of all yarn center positions throughout the scanned volume.

\subsection{Semi-automatic annotations}

\begin{figure}[htbp]
    \centering
    \includegraphics[width=0.8\textwidth]{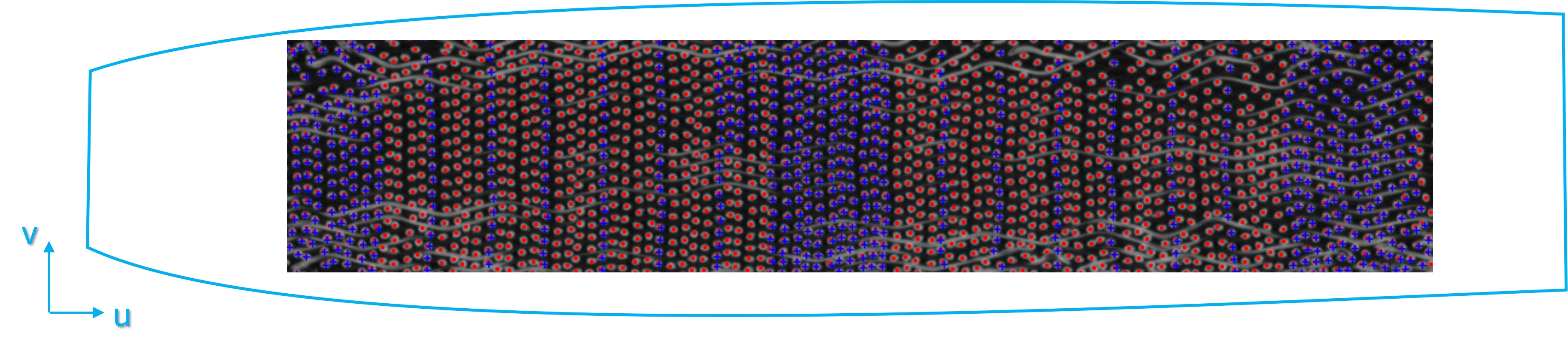}
    \caption{Blue crosses: Manual annotations made by experts. Red: Additional semi-automatically annotated planes. Cyan lines: theoretical contours of the part. ($w=500$-plane)}
    \label{fig:annotations_full}
\end{figure}

In the tomographic images, the approach expressed in~\Cref{subsec:yarn_detection} makes it possible to identify more yarns than those covered by the available, sparse annotations (\Cref{fig:annotations_full}). A few yarns remain undetected, typically due to their atypical shapes or close proximity to neighboring yarns.
These cases are manually annotated, making the overall procedure semi-automatic.
A necessary trimming is applied to the preform in order to shape the blade, as a result, many yarns are cut, causing their number to decrease when moving toward the tip of the blade.

After the procedure, our main database can be expanded by producing a complete annotation of four slices located at planes 0, 500, 1000, and 1500, containing 3057, 2682, 2017 and 1970 annotated yarns, respectively.
In these slices, we manually added 197, 149, 189, and 277 yarns, which corresponds to respectively 6.5\%, 5.6\%, 9.4\%, and 14.1\% of the total yarns in each slice. Towards the top of the part, the weaving is less rigid, which causes yarn columns to merge and complicates the automatic detection task, thus increasing the number of manually annotated yarns.
These annotations serve later as initialization and validation points (\Cref{fig:annotations_full}).

\section{Characterization and modeling of material features}
\label{sec:characterization}

The primary objective of this work is to develop a general approach for extracting continuous yarn paths from large tomographic scans of textile composite preforms.
Rather than focusing on segmentation, the method aims at reconstructing the full three-dimensional trajectories of yarns by determining their positions in successive tomographic planes.
This process is demonstrated on a fan blade preform, but it is designed to be applicable to a wide range of woven composite architectures.

In each 2D slice, yarn positions are identified and globally analyzed to ensure global spatial consistency.
Since yarn \textbf{cross-sections} can exhibit complex and irregular shapes that deviate from ideal ellipses, the method incorporates a statistical characterization of these cross-sectional geometries derived from Principal Component Analysis (PCA) of annotated data.
This characterization allows for a generalized, data-driven description of cross-section variability across the fabric.

Beyond the slice-wise analysis, the method also models the \textbf{continuity} of each yarn along its length.
The evolution of yarn positions through successive planes is used to regularize the reconstruction, ensuring smooth and physically consistent paths throughout the volume.
Additionally, local spatial \textbf{interactions}  between neighboring yarns are explicitly modeled to preserve the structural organization of the weave. By integrating these elements, all derived from real tomographic data and manual annotations, a variational formulation is provided. The objective is to simultaneously segment all yarns of a single type, (warp or weft) in each plane while maintaining consistent yarn identities across the volume.

All annotations were used to ensure spatial representativeness, though fewer may suffice if they remain well distributed. A minimal annotation analysis is provided in \Cref{sec:appendix_study}. Computations were implemented in Python using \texttt{NumPy}, \texttt{SciPy} (including SVD for PCA), and \texttt{scikit-learn} for GMM estimation.

\subsection{Yarn cross-section}
\label{subsec:yarn_detection}

Based on the annotated central coordinates of selected yarns, a bounding box of fixed size $h\times w$ is defined around each point to extract localized window.
Each patch captures the local grayscale information around the yarn axis and is vectorized into a row vector of length $q = h\times w$.
All such vectors are then assembled into a data matrix $\textbf{M} \in \mathbb{R}^{p \times q}$ where $p$ is the number of annotations.

\begin{figure}[h!]
\centering
\begin{subfigure}{0.48\textwidth}
    \centering
    \includegraphics[width=\linewidth]{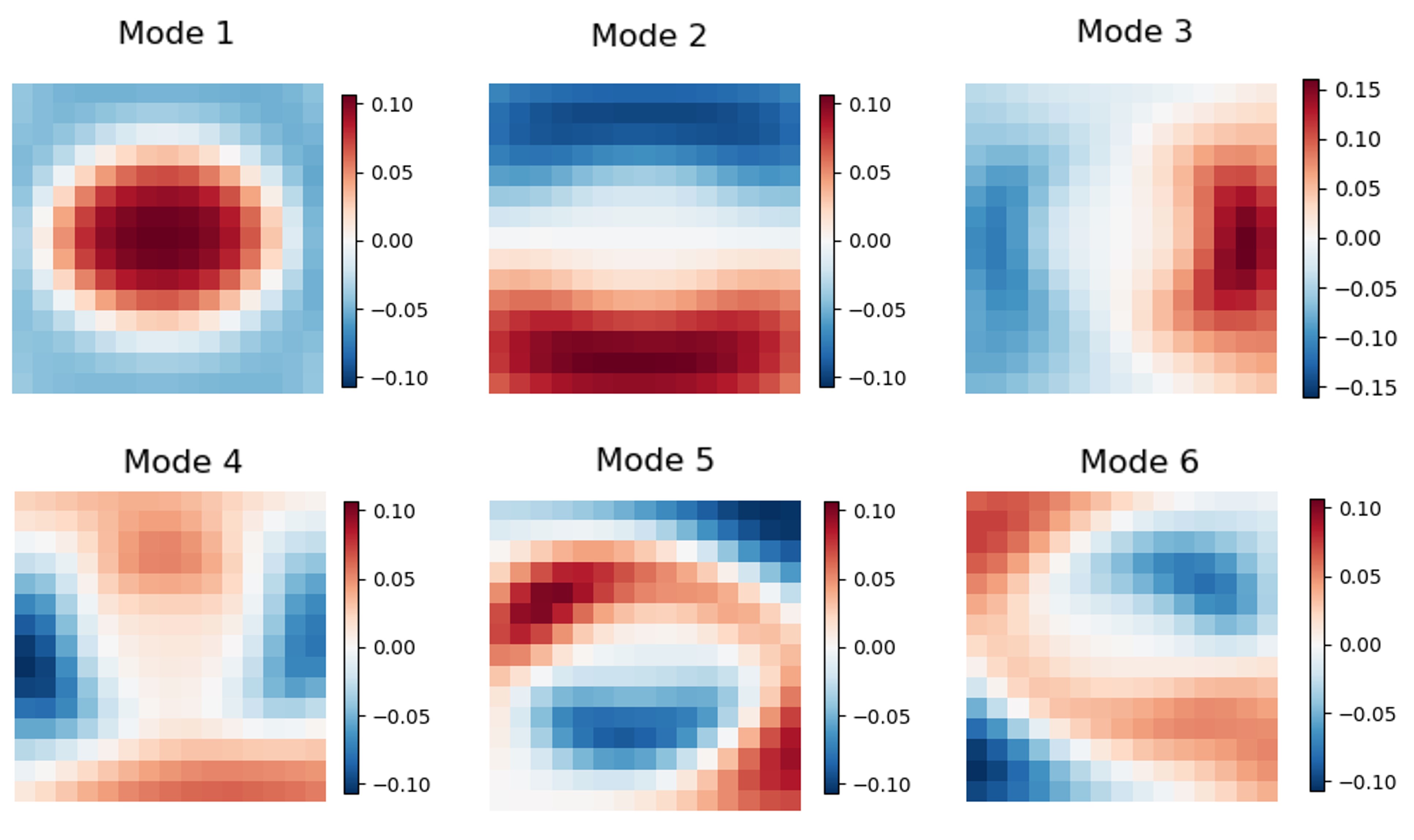}
    \caption{Visualization of the six first PCA modes for warp yarns.}
    \label{fig:modes_pca}
\end{subfigure}
\hfill
\begin{subfigure}{0.48\textwidth}
    \centering
    \includegraphics[width=\linewidth]{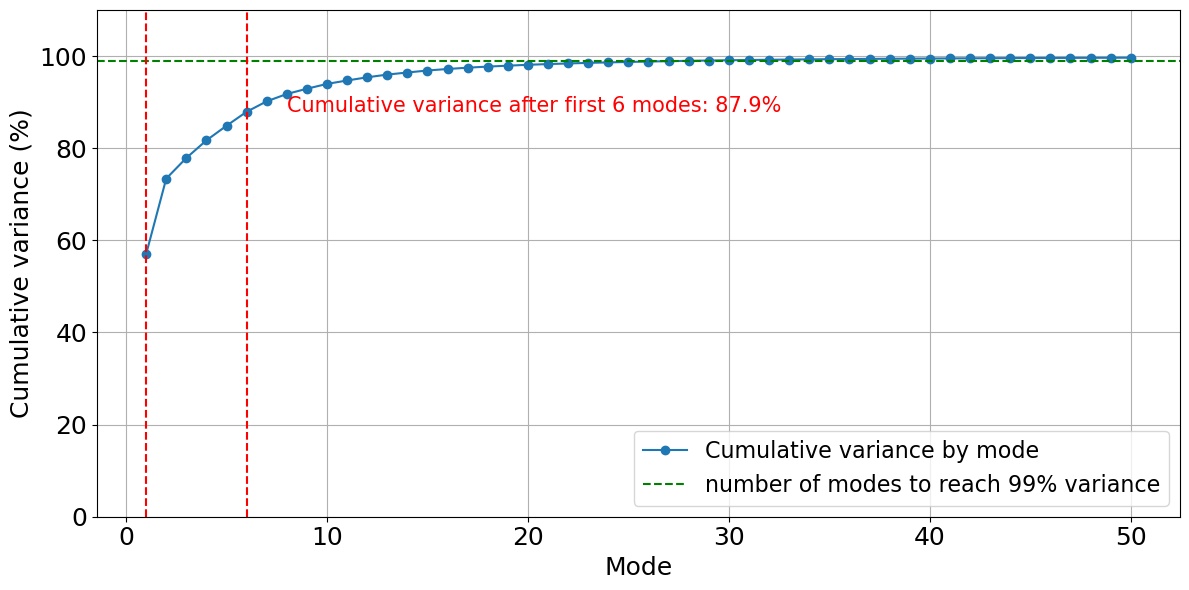}
    \caption{Variance analysis per mode via PCA.}
    \label{fig:pca_eigenvalues}
\end{subfigure}

\caption{PCA analysis of warp yarns: mode visualization and associated variance.}
\label{fig:pca_combined}
\end{figure}

To analyze and interpret the main modes of variation across this dataset, a Principal Component Analysis (PCA) is applied to the matrix $\mathbf{M}$.
PCA is well suited in this context as it provides a compact and interpretable decomposition of the dataset into orthogonal directions of progressively decreasing importance to take into account the yarn appearance across all samples.
This decomposition results in a set of orthogonal principal components $\mathbf V_i$, ordered by decreasing eigenvalues $\lambda_i$, which represent the dominant modes of variation within the dataset. Each mode can be reshaped back into an image of the original patch size, allowing for direct visual interpretation. Examination of the PCA cumulative explained variance (\Cref{fig:pca_eigenvalues}), obtained from the eigenvalues of the covariance matrix, indicates that the first six modes capture nearly 90\% of the total variance.

Before applying PCA, all intensity patches are centered to remove global offsets, ensuring that the principal modes capture the main patterns of variation.
The first principal mode represents the average cross-section image aligned with the yarn axis while higher-order modes (\Cref{fig:modes_pca}) reflect specific types of variation, including small translations (vertical and horizontal shifts), local rotations of the yarn within the bounding box, and changes in cross-section shape, such as flattening. Together, these modes provide a low-dimensional yet physically meaningful description of the diversity of yarn cross-sections observed in the data.

Since our study focuses on yarn-level annotations of warp yarns, the first mode --- also called ``fundamental'' in the following --- which encapsulates the most significant geometric and textural pattern, is the most relevant for tracking.

Modes 2 and 3 exhibit a strong proximity to the two components of the gradient of mode 1. This indicates that their primary function is to compensate for inaccuracies in the annotated yarn center, and thus in the extraction window, which are expected due to the manual annotation procedure. Since we represent yarns through their evolving centerline in a continuous domain, displacements are handled directly via the position of the center point and the associated shape. In this setting, the contribution of gradient-like modes is effectively embedded in the trajectory of the centerline and does not require explicit modeling.

\begin{figure}[ht!]
    \centering

    \begin{subfigure}[b]{0.2\textwidth}
        \includegraphics[width=1.5\textwidth]{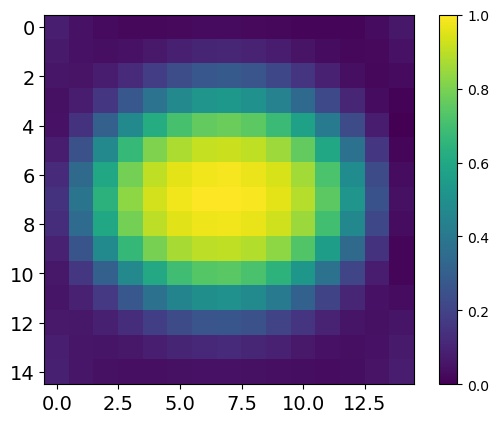}
        \caption{}
    \end{subfigure}
    \hspace{0.5cm}
    \begin{subfigure}[b]{0.33\textwidth}
        \includegraphics[width=\textwidth]{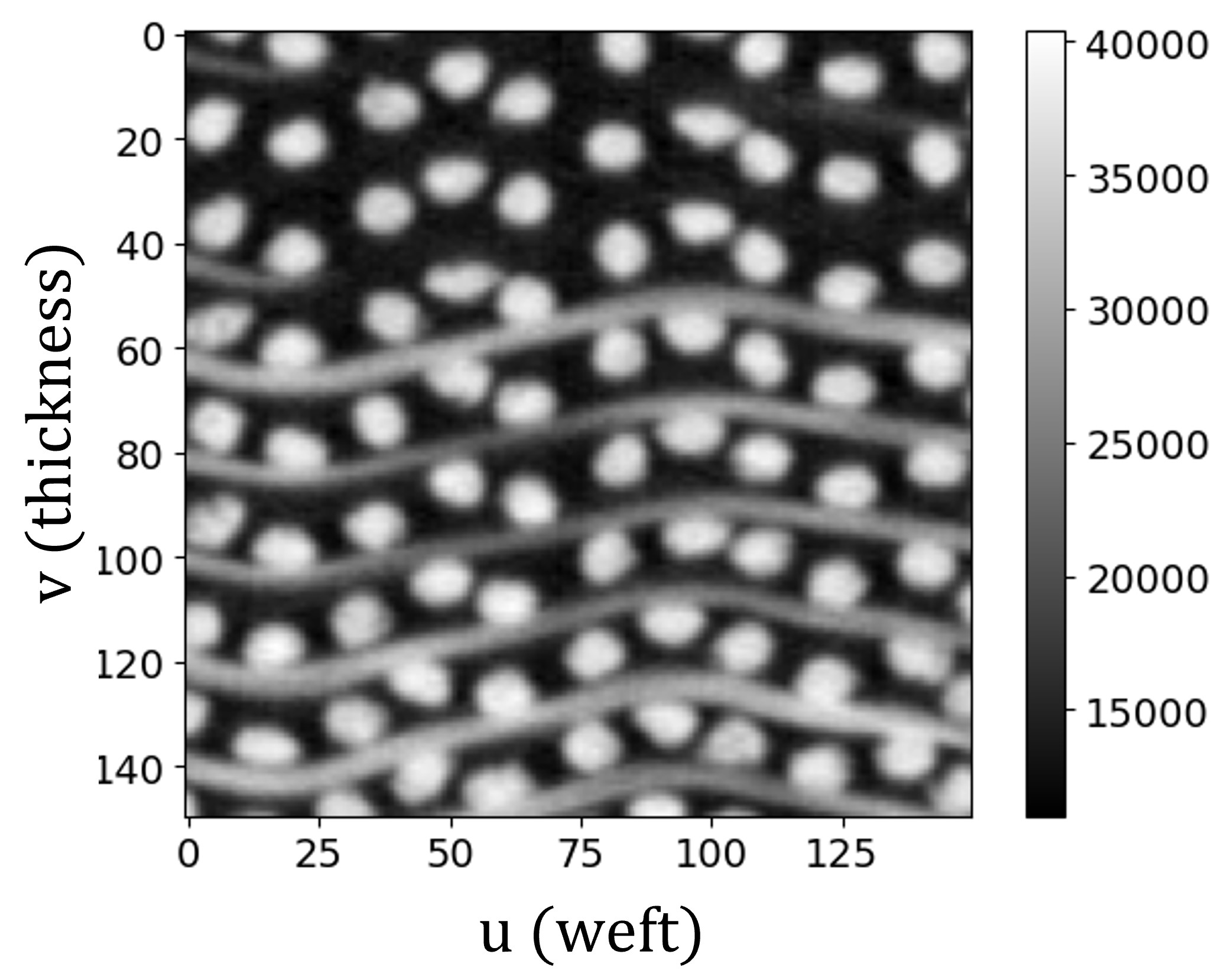}
        \caption{}
    \end{subfigure}
    \hspace{0.5cm}
    \begin{subfigure}[b]{0.33\textwidth}
        \includegraphics[width=\textwidth]{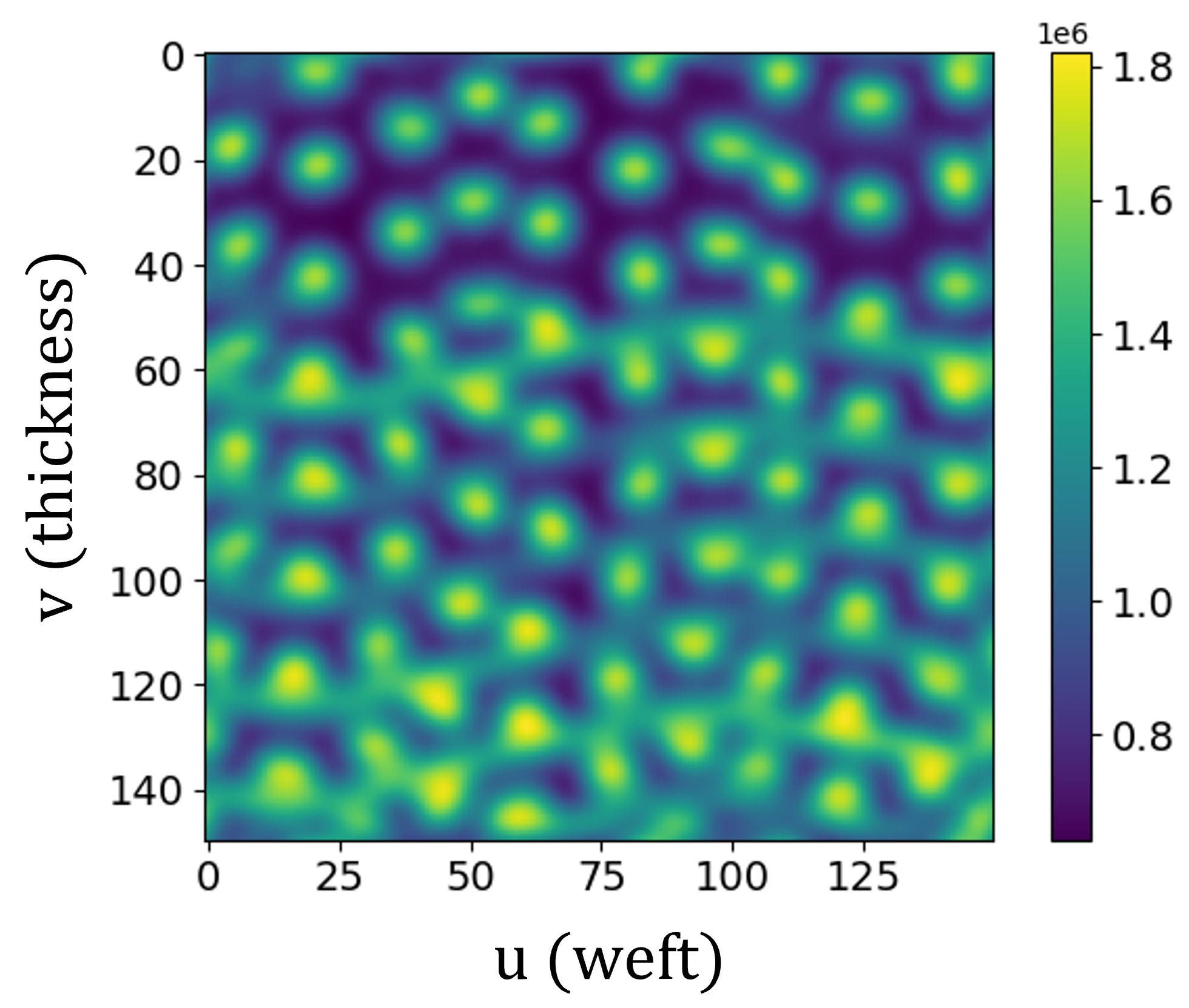}
        \caption{}
    \end{subfigure}

    \caption{(a) The fundamental mode $\phi$, (b) 2D slice of CT with warp yarn cross-sections, (c) Cross-correlation result.}
    \label{fig:cross-correlation}
\end{figure}

Only the fundamental mode $\phi$ is retained, while all other modes (describing rotation, flattening, $\cdots$) are discarded. For visualization purposes, the fundamental mode is normalized between 0 and 1. This fundamental mode is then used to generate ``heatmaps'' $P_1$ by cross-correlating it with the tomographic volume slices $I_w$. Because $\phi$ is centrosymmetric $\phi(-\mathbf X)=\phi(\mathbf X)$, the cross-correlation is obtained by a mere convolution between $\phi$ and $I_w$.
\begin{equation}
J(\mathbf X) = \big( I_w * \phi \big)(\mathbf X)
\end{equation}
In practice, the mode $\phi$ is zero-padded prior to the operation so that the resulting heatmap $P_1$ has the same spatial dimensions as the slice $I_w$.
Such a map (\Cref{fig:cross-correlation}) approximates the likelihood that a yarn center is located at each considered point. Consequently, local maxima appear at the center of each yarn, where the cross-section matches the mode as closely as possible and the heatmap is simply read at the position of yarn centers
\begin{equation}
P_1\big(\mathbf{X}_i(w)\big) = J(\textbf{X}_i(w)).
\end{equation}

\subsection{Yarn continuity}

\begin{figure}[ht!]
  \centering
  \begin{subfigure}[b]{0.30\linewidth}
    \includegraphics[width=\linewidth]{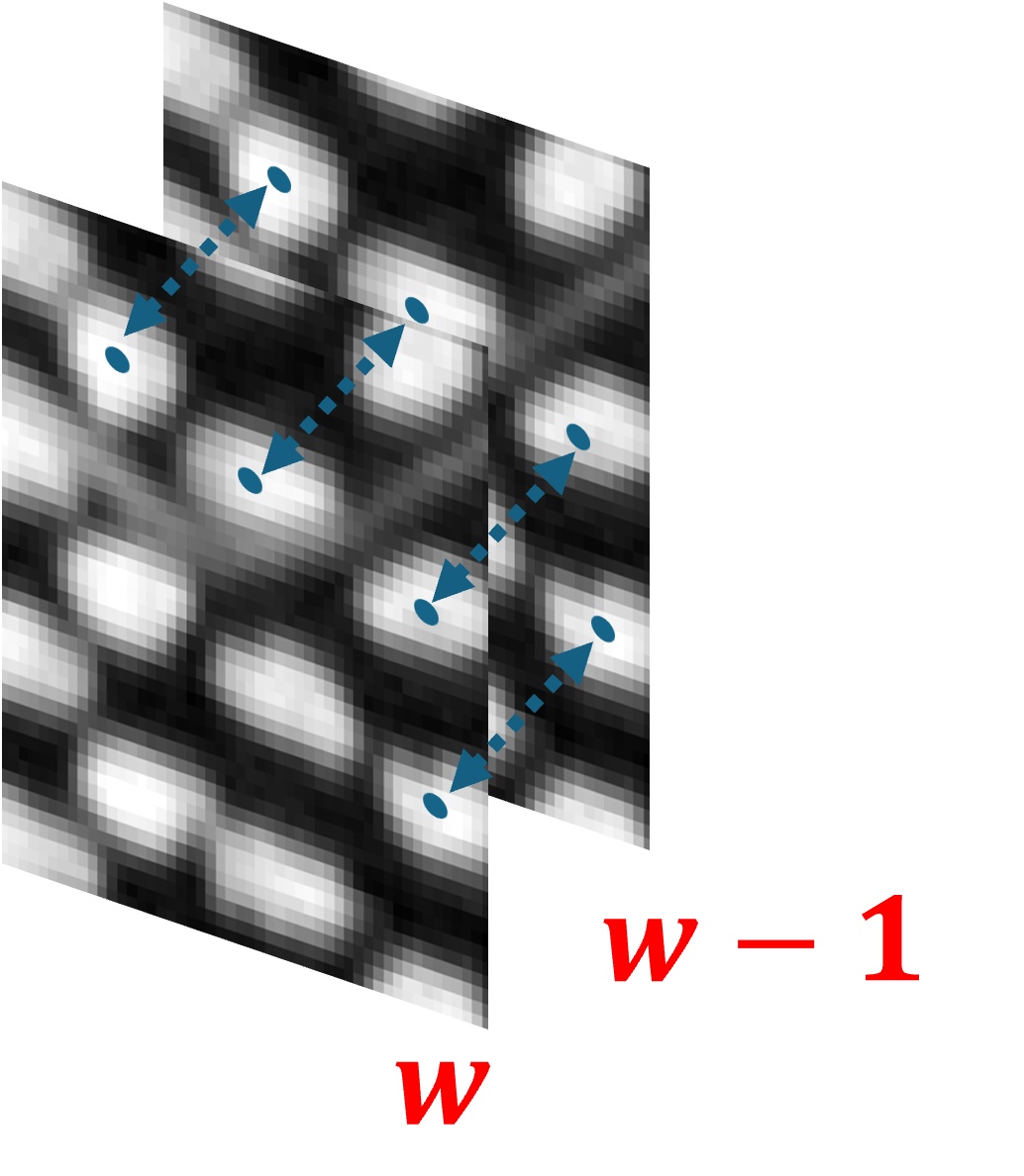}
    \caption{}
    \label{fig:tracking_scheme}
  \end{subfigure}
  \hspace{0.5cm}
  \begin{subfigure}[b]{0.30\linewidth}
    \center
    \includegraphics[width=1.2\linewidth]{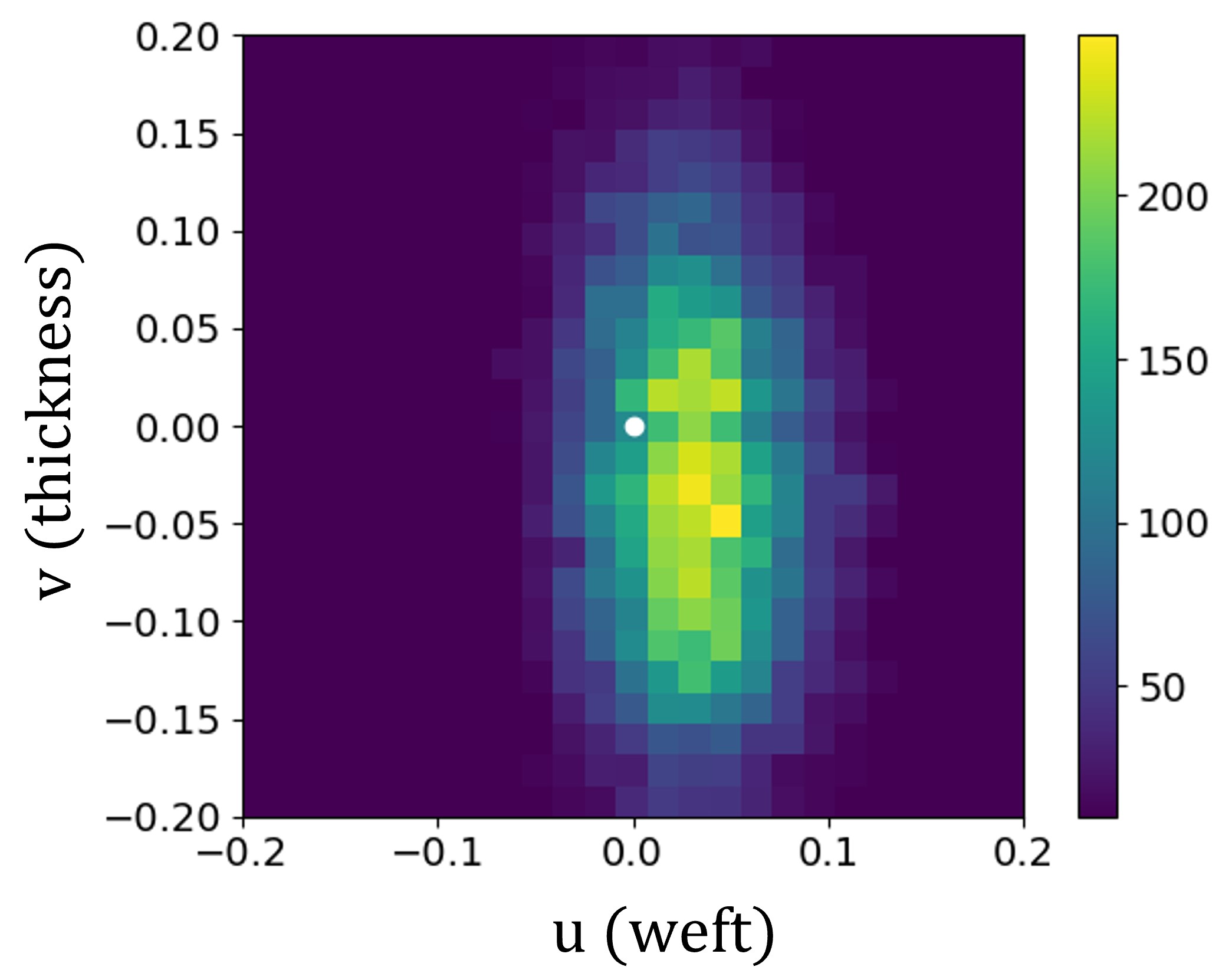}
    \caption{}
    \label{fig:tracking_histogram}
  \end{subfigure}
  \caption{(a) Yarn misalignment, along their paths (blue arrows); (b) Histogram of the displacement vector over consecutive planes. \\
  }
\end{figure}

It is essential to preserve the continuity and smooth variation of the yarn path and shape. To characterize this regularity, let us define the ``displacement'' vector $\bm{X}_i(w)-\bm{X}_i(w-1)$ as the yarn path increment, projected onto a cross-sectional plane, where $\bm{X}_i(w)$ denotes the 2D position of yarn $i$ at tomographic plane $w$ (\Cref{fig:tracking_scheme}).

From the data, the statistical distribution of displacement vectors,
$P_2\big(\mathbf{X}_i(w) - \mathbf{X}_i(w-1)\big)$, can be computed (\Cref{fig:tracking_histogram}).

A modeling of this histogram is provided by a Gaussian Mixture Model (GMM):
\begin{equation}
P_2\big(\mathbf{X}_i(w) - \mathbf{X}_i(w-1)\big) =
\sum_{k=1}^{K} \varpi_{2_k} \, \mathcal{N}(\bm{\mu}_{2_k}, \bm{\Sigma}_{2_k}),
\end{equation}
where \( \varpi_{2_k} \) are the mixture weights, and
\( \mathcal{N}(\bm{\mu}_{2_k}, \bm{\Sigma}_{2_k}) \) denotes a 2D normal distribution
with mean displacement vector \( \bm{\mu}_{2_k} \) and covariance matrix \( \bm{\Sigma}_{2_k} \).
The corresponding probability density function is given by
\be
\mathcal{N}(\bm{\mu},\bm{\Sigma}) =
(2\pi)^{-1} |\bm{\Sigma}|^{-1/2} \exp\!\left(-\tfrac{1}{2}(\mathbf{x}-\bm{\mu})^{\top}\bm{\Sigma}^{-1}(\mathbf{x}-\bm{\mu})\right).
\ee

This formulation provides a probabilistic description of yarn displacements between consecutive planes, serving as a statistical prior that enforces the observed regularity of yarn trajectories throughout the volume.

In practice, the observed distribution of displacements is well captured by a single Gaussian component (\(K = 1\)).
In this case, the mixture model simplifies to a single normal distribution
\begin{equation}
\label{eq:P2}
P_2\big(\mathbf{X}_i(w) - \mathbf{X}_i(w-1)\big) = \mathcal{N}(\bm{\mu_2}, \bm{\Sigma_2}),
\end{equation}
which corresponds to assuming that the yarn displacements follow a 2D normal law with mean vector \(\bm{\mu_2} \in \mathbb{R}^{2}\) and covariance matrix  \(\bm{\Sigma_2} \in \mathbb{R}^{2\times2}\).

The distribution shown in~\Cref{fig:tracking_histogram} is coherent with our expectations.
Since the fabric is woven by columns, the yarns exhibit a predominantly vertical motion, which explains the elongated distribution along the vertical axis.
The mean is not exactly centered at $(0, 0)$, likely due to a slight offset in the positioning of the specimen within the tomograph, which induces a small global drift of the trajectories.
The observed inter-slice displacement of approximately $\pm 0.15$ is also consistent with the characteristics of this fabric, which is tightly woven and well organized, leading to very limited motion between consecutive slices.

\subsection{Yarn interactions}

\begin{figure}[ht!]
  \centering
  \begin{subfigure}[b]{0.30\linewidth}
    \center
    \includegraphics [width=\linewidth]{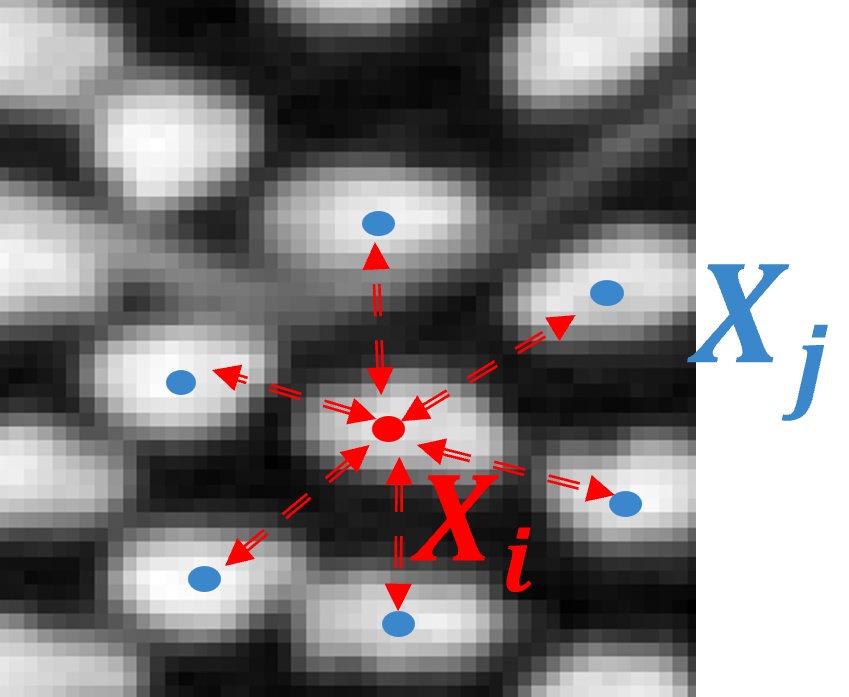}
    \caption{}
    \label{fig:neighbors_scheme}
  \end{subfigure}
  \hfil
  \begin{subfigure}[b]{0.30\linewidth}
    \center
    \includegraphics[width=1.2\linewidth]{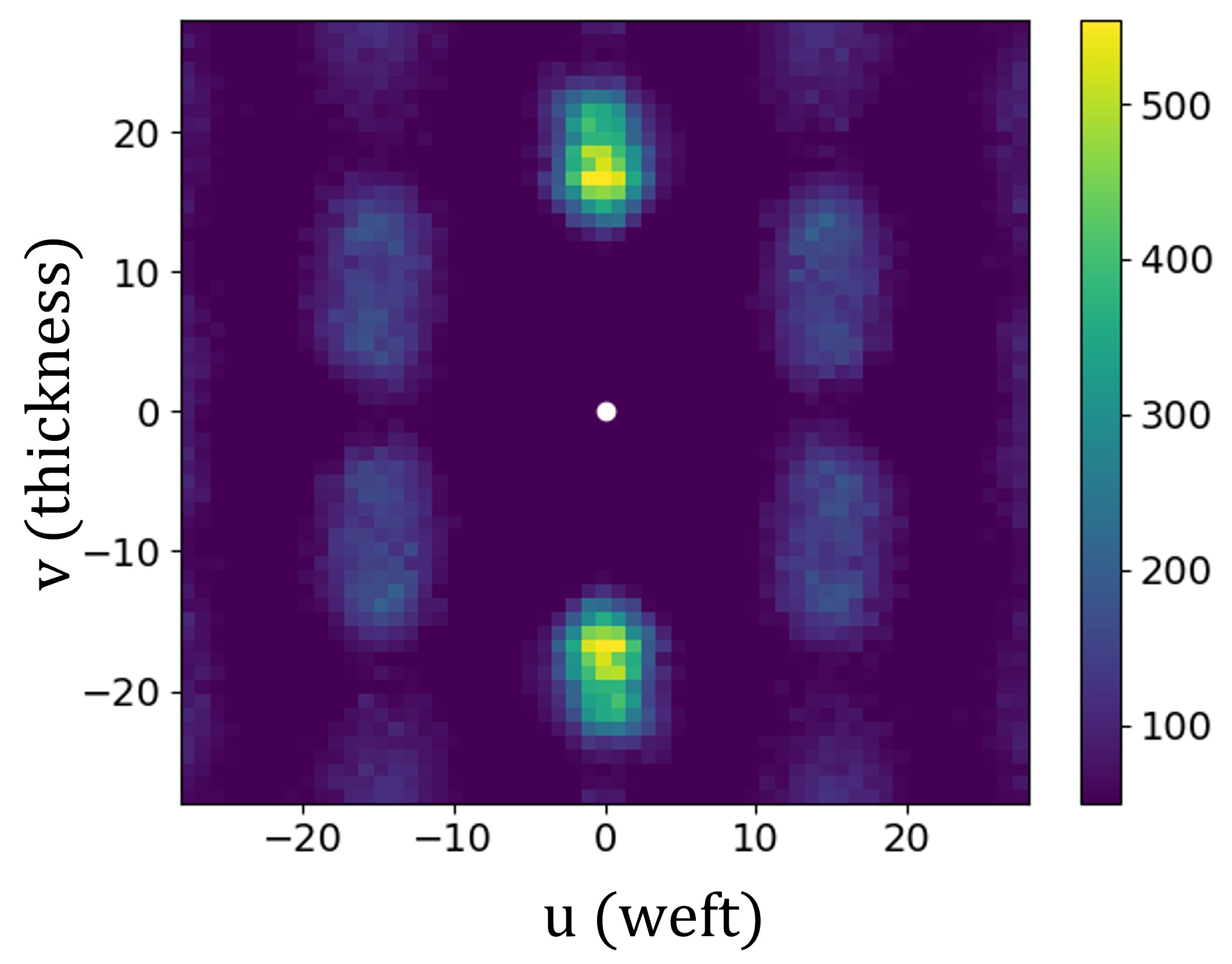}
    \caption{}
    \label{fig:neighbors_histogram}
  \end{subfigure}
  \caption{
  (a) Yarn arrangement with respect to its neighbors (red arrows); (b) Histogram of the intercenter vectors with respect to a central yarn. \\
  }
\end{figure}

To characterize the interactions between a yarn and its neighbors, inter-center vectors $\bm X_i(w) - \bm X_j(w)$ are computed for each pair of points $\bm X_i(w), \bm X_j(w)$ (see~\Cref{fig:neighbors_scheme}). Only vectors with a norm less than or equal to a specified radius $r$ are retained for further analysis. The radius $r$ is chosen to include only the first set of neighbors to each yarn center. \(\mathcal{V}_i\) is the set containing those neighbors.
\Cref{fig:neighbors_histogram} presents the 2D histogram of all collected inter-center vectors for the analyzed warp yarns.
This distribution reveals an hexagonal arrangement of neighboring yarns.

This choice to model the yarn environment locally, rather than globally, comes from the repetitive and periodic nature of the weaving process.
The structure of the textile is such that the local configuration is representative of the overall pattern.
Higher-order relationships reflect the same arrangement due to the regularity of the weave, making their explicit modeling redundant.
Since our approach treats each yarn individually and focuses on the immediate neighbors, capturing only the first neighbors is sufficient to describe the local environment accurately.

Again, this distribution can be modeled using a Gaussian Mixture Model (GMM) with $|\mathcal{V}_i|$ 2D Gaussian components.
Here, $|\mathcal{V}_i|=6$, correspond to the first neighbors of a central warp yarn $i$.
The observed arrangement of neighboring yarns reflects the underlying manufacturing process directly.
Indeed, as shown in~\Cref{fig:neighbors_histogram}, the inter-column distance remains almost constant, consistent with the regular spacing imposed by the loom.
Moreover, the positions of yarns within the same column are more precisely defined than those of yarns in adjacent columns.
This anisotropy arises from the dominant vertical motion of yarn columns during weaving, which constrains their relative alignment.
Conversely, the positions of yarns in neighboring columns exhibit greater variability, a consequence of the weave pattern itself, whose periodicity manifests over several columns rather than a single neighboring relation.

Note that, when computing the distance between a yarn $i$ and its first neighbors, the local statistics of each neighbor are determined by its own Gaussian component within the GMM.
Therefore, $\boldsymbol{\Sigma}_{3_j}$ $\in \mathbb{R}^{2\times2}$ represents the covariance matrix of the Gaussian component assigned to point $j$, and similarly, $\boldsymbol{\mu}_{3_j}$ $\in \mathbb{R}^{2}$ denotes its corresponding mean.
The probability is thus evaluated with respect to the Gaussian distribution associated with each neighbor component. Which leads to
\be
\label{eq:P3}
P_3(\bm X_i - \bm X_j) = \mathcal{N}(\bm{\mu}_{3_j}, \bm{\Sigma}_{3_j})
\ee

\section{Tracking strategy}
\label{sec:strategy}

The extraction of yarn paths is formulated as a tracking procedure employing a variational formulation composed of three energy terms, \(E_1\), \(E_2\), and \(E_3\), derived from the properties introduced in \Cref{sec:characterization}.
Here, we briefly recall their formulation to motivate the optimization derivations that follow.
The first term, $E_1\equiv -P_1$, encodes the local image-based alignment and is minimized when all yarn center points \(\mathbf{X}_i(w)\) coincide with local maxima of the heatmap $P_1$, thus driving accurate localization within each tomographic slice.
The second term, $E_2$, promotes longitudinal smoothness and geometric coherence across slices.
It can be interpreted as minimizing the negative log-likelihood of observing the current yarn trajectories under a smoothness prior, thereby favoring physically plausible paths.
Similarly, $E_3$ introduces a global structural prior that captures the crystalline-like arrangement of the woven architecture.
Minimizing this global energy enforces consistency between the reconstructed paths and the expected general pattern.
In all cases, the energies are aggregated over all yarns and slices, ensuring a coherent balance between local fidelity, regularity, and structural consistency in the optimization process.

\subsection{Variational formulation}

Each warp yarn \( i \) is represented by a parametric function $\mathbf{X}_i : \mathbb{R} \rightarrow \mathbb{R}^2$ , which defines the position of its center in the \((u,v)\) cross-sectional plane at slice coordinate \( w \).
The global configuration of all \( N_t \) yarns at a given slice is denoted $\mathbf{X}(w) = \{ \mathbf{X}_i(w) \}_{i=1,\dots,N_t}$. Starting from the initial configuration at \( w = w_0 \), the objective is to iteratively compute \(\mathbf{X}(w)\) from \(\mathbf{X}(w-1)\), reconstructing the full yarn system across the volume.

\subsubsection{Optimization scheme}

The reconstruction is cast as the minimization of a global energy functional
\[
E(\mathbf{X}(w)) = E_1(\mathbf{X}(w)) + \alpha\, E_2(\mathbf{X}(w)\,|\,\mathbf{X}(w-1)) + \beta\, E_3(\mathbf{X}(w)),
\]
where \(\alpha\) and \(\beta\) are scalar weighting coefficients.

The global energy $E$ is smooth but non-convex, and may exhibit multiple local minima due to the periodic yarn organization and Gaussian-mixture interaction term. However, since the three contributions are linearly combined, we expect a single minimum near the desired solution.

The weights $\alpha$ and $\beta$ were determined empirically. After normalizing the energy terms, a sensitivity analysis was performed by progressively adding $E_2$ and $E_3$ to $E_1$, adjusting $\alpha$ and $\beta$ to ensure balanced contributions without dominance.

\noindent The total gradient and Hessian are assembled as

\vspace{-0.5cm}
\begin{align}
\boldsymbol{\nabla} E &= \boldsymbol{\nabla} E_1 + \alpha\, \boldsymbol{\nabla} E_2 + \beta\, \boldsymbol{\nabla} E_3, \\
\mathbf{K} &= \boldsymbol{\nabla}\boldsymbol{\nabla}^\top E = \mathbf{H}_1 + \alpha\, \mathbf{H}_2 + \beta\, \mathbf{H}_3.
\end{align}

Then, minimization of \(E\) is performed using an iterative Gauss-Newton procedure. At each iteration, the linearized system
\begin{equation}
\mathbf{K}\, \delta \boldsymbol{X} = - \boldsymbol{\nabla} E
\end{equation}
is solved to compute the incremental update \(\delta \boldsymbol{X}\).
The matrix $\mathbf{K} \in \mathbb{R}^{2N_t \times 2N_t}$ corresponds to the global Hessian of the energy, where \(N_t\) is the number of yarn centers in the current slice (each center being represented by its two in-plane coordinates).
The vector $\delta \boldsymbol{X} \in \mathbb{R}^{2N_t \times 1}$ contains the displacement increments for all yarn centers.
The configuration is then updated as $\mathbf{X_i}(w) = \mathbf{X_i}(w-1) + \delta \boldsymbol{X_i}$.

This is performed iteratively until all slices have been processed.

To accelerate the optimization of the likelihood-based terms, we introduce a Potts framework, where the assignment of each point to a Gaussian component is represented as a discrete variable.

\subsubsection{Energy terms}

In the following, we detail the required energies, their gradients and Hessians to compute the optimization.  The set of unknowns $\mathbf X_i(w)$ is represented as a $2\times N_t$ vector $(u_1,v_1,u_2,v_2,..., u_{N_t}, v_{N_t})^\top$.
$E_1=-P_1(\mathbf{X}_i(w))$ is the opposite of the heatmap value summed over all yarn centers, $i$. The negative sign in $E_1$ reflects that heatmap maxima correspond to yarn centers, while the problem is posed as a minimization.

\vspace{-0.5cm}
\begin{align}
E_1(\mathbf{X}(w)) &= -\sum_i J(\mathbf{X}_i(w)) \\
E_2(\mathbf{X}(w)\,|\,\mathbf{X}(w-1)) &= - \sum_i \log P_2(\mathbf{X}_i(w) - \mathbf{X}_i(w-1)). \\
E_3\big(\mathbf{X}(w)\big) &= - \sum_i \sum_{j \in \mathcal{V}_i}
\log P_3\Big( \mathbf{X}_i(w) - \mathbf{X}_j(w) \Big)
\end{align}

 The set of first and second derivatives of $E_1$ for all variations of the yarn centers are needed.
 It is convenient for this purpose to compute the derivatives from the convolution of $I_w$ with $(\partial_u\phi,\partial_v \phi, \partial^2_{uu}\phi, \partial^2_{uv}\phi, \partial^2_{vv}\phi)$ to produce $(\partial_u J,\partial_v J, \partial^2_{uu}J, \partial^2_{uv}J, \partial^2_{vv}J)$.
 The sampling of the first two scalar fields at the $N_t$ yarn positions $\mathbf X_i$ yields a $(2\times N_t)$ vector $\bm \nabla E_1$,
\be\ba{rcl}
\bm \nabla E_1 &=& (\partial_{u_1} E_1,\partial_{v_1} E_1, .... \partial_{u_{N_t}} E_1,\partial_{v_{N_t}} E_1)^\top\\
&=& (-J_{u}(\mathbf X_1), -J_{v}(\mathbf X_1), -J_{u}(\mathbf X_2), -J_{v}(\mathbf X_2), ... -J_{u}(\mathbf X_{N_t}), -J_{v}(\mathbf X_{N_t}))^\top
\ea\ee
Similarly, the Hessian $\mathbf H_1$ is constructed as a block diagonal matrix, composed of the negative of the successive $2\times 2$ curvature tensors $\bm \kappa_i$, sampled from the three continuous fields $\partial^2_{uu}J, \partial^2_{uv}J, \partial^2_{vv}J)$ at the successive yarn centers $\mathbf X_i(w)$:
\be
\bm \kappa_i=\left(\ba{cc}
\partial^2_{uu}J(\mathbf X_i) &\partial^2_{uv}J(\mathbf X_i)\\
\partial^2_{uv}J(\mathbf X_i) &\partial^2_{vv}J(\mathbf X_i)\\
\ea\right).
\ee
These $\kappa_i$ for $i=1, .. N_t$ are assembled along the diagonal of $\mathbf H_1$.

For the second energy term, its gradient is the concatenation of 2d-vectors, one for each yarn $i=1,...,N_t$. These elementary vectors  $(\partial_{u_i}E_2,\partial_{v_i}E_2)$ are equal to the product of the inverse covariance matrix, $\boldsymbol \Sigma_2^{-1}$, with the incremental displacement corrected by the mean, $(\mathbf X_i(w)-\mathbf X_i(w-1)-\bm \mu_2)$. The Hessian $\mathbf H_2$ is again a block diagonal matrix composed of $N_t$ replicas of the $2\times 2$ matrix $\boldsymbol\Sigma_2^{-1}$.

Finally, for $E_3$, its gradient is again the concatenation of $i=1,...,N_t$ 2d-vectors, given by
\be
(\partial_{u_i}E_3,\partial_{v_i}E_3)^\top =
\sum_{j \in \mathcal{V}_i} \boldsymbol{\Sigma}_{3_j}^{-1} \Big( (\mathbf{X}_i(w) - \mathbf{X}_j(w)) - \boldsymbol{\mu}_{j} \Big)
\ee

The Hessian matrix $\mathbf{H}_3 \in \mathbb{R}^{2N_t \times 2N_t}$ is a symmetric block matrix, each $2 \times 2$ block $\mathbf h_3(i,j)$ is defined as
\be
\mathbf{h_3}(i,j) =
\begin{cases}
-\boldsymbol{\Sigma}_{3_j}^{-1}, & \text{if }~ (i \neq j) \& (j \in \mathcal{V}_i), \\
\displaystyle \sum_{j \in \mathcal{V}_i} \boldsymbol{\Sigma}_{3_j}^{-1}, & \text{if } (i = j), \\
0, & \text{if } j \notin \mathcal{V}_i.
\end{cases}
\ee

\subsection{Initialization}

The tracking of warp yarns was initiated from the annotated slice, located at the blade root ($w = 0$), and propagated upwards through 1500 tomographic planes, following a total of $N_t = 3057$ warp yarns.
This choice is primarily motivated by the fact that the tracking framework can handle yarns that gradually disappear from view (yarn exits) but not the inverse case, where new yarns suddenly appear. The algorithm determines the boundaries of the part, and based on their position, it terminates the tracking of any yarn that extends beyond them.
Starting from the root region therefore ensures that all warp yarns are initially visible, maximizing the number that can be successfully tracked.
Moreover, the root area exhibits a denser and more compact weave, as this region is designed to withstand higher mechanical loads.
The reduced yarn mobility in this zone contributes to more stable and reliable initial extraction.

Because the yarn cross-sections exhibit an anisotropic, approximately elliptical shape, the notion of distance needs to take this into account. The local coordinate system around each yarn center is transformed into an anisotropic reference frame through a mapping function $\mathbf{A}:~\mathbb{R}^2 \rightarrow \mathbb{R}^2$.
This transformation maps the elliptical geometry into a circular one, allowing the definition of an anisotropic distance metric that properly scales the coordinates along the principal yarn directions.

The parameters of the transformation $\mathbf{A}$, specifically the lengths of the ellipse axes $a$ and $b$, are directly inferred from the fundamental mode obtained through PCA. During initialization, this anisotropic metric identifies unrealistically close yarn centers and separates them to enforce physically consistent positions, ensuring reliable proximity evaluation and preventing incorrect associations between neighboring yarns.

\section{Evaluation metrics}
\label{sec:metrics}
\subsection{Instance detection per slice}
\label{subsec:slice_tracking}

To quantify tracking errors, a tolerance domain is defined around each reference yarn position, modeled as an ellipse reflecting the anisotropic yarn cross-section and its mean dimensions. Denoting the reference and predicted positions of yarn $i$ at slice $w$ by $\mathbf{X}_{\text{ref},i}$ and $\mathbf{X}_{\text{pred},i}$, the tracking indicator is defined as
\begin{equation}
\mathrm{C}_i(w) = 1 \;\text{if}\;
\left\| \mathbf{A}(\mathbf{X}_{\text{pred},i} - \mathbf{X}_{\text{ref},i}) \right\| \le 1,\ \text{else } 0
\end{equation}
where $\mathbf{A}$ denotes the anisotropic transformation associated with the tolerance ellipse.

To assess robustness, standard detection metrics are used: a predicted yarn is a true positive (TP) if it lies within a spatial threshold of an annotation, while missing annotations define false negatives (FN) and unmatched predictions define false positives (FP). From these, sensitivity, precision, accuracy, and F1-score are computed. All results are summarized in~\Cref{tab:yarn_metrics}.

\subsection{Aggregated yarn paths}
\label{subsec:aggreg_tracking}

To evaluate the 3D tracking performance of the algorithm, a correspondence is established between the dense annotations generated for the initialization slice and the available manual annotations.
In the initialization slice, all yarn sections are annotated, resulting in a dense set of $3057$ yarns, whereas the manual annotations cover only selected regions and one out of every five columns elsewhere, yielding $1170$ annotated yarns. In both sets, each yarn is assigned an identifier, allowing its tracking and identification in the 3D volume. A one-to-one mapping between the $1170$ yarns present in both annotation sets is obtained by matching their identifiers.
This enables a comparison of their trajectories across the volume.
As manual annotations are available every $35$ slices, the tracking accuracy is evaluated at each annotated slice along the trajectory.

For each yarn $i$, the reference annotations $\mathbf{X}_{\text{ref},i}(w)$ and the corresponding predicted positions $\mathbf{X}_{\text{pred},i}(w)$ are compared at the annotated slice positions $w$.
Tracking correctness at each slice is evaluated using the same anisotropic criterion defined in the previous subsection, ensuring consistency with the local per-slice metric.

To extend this evaluation to entire trajectories, we define a cumulative validity indicator for each yarn as

\begin{equation}
V_i(w) = 1 \;\text{if}\;
\bigcup_{\alpha=1}^{w}
\mathbf{A}\!\left(\mathbf{X}_{\text{pred},i}(\alpha)-\mathbf{X}_{\text{ref},i}(\alpha)\right)
\le 0,\ \text{else } 0
\label{eq:cumulative_validity}
\end{equation}
This indicator takes the value $1$ if, up to slice $w$, the predicted trajectory of yarn $i$ remains within the anisotropic tolerance domain at every intermediate slice.  The sum over all yarns $i$ of $V_i$ provides a cumulative tracking metric, $V(w)$.

Once a mismatch occurs at a given slice, the trajectory is considered invalid from that point onward.
To characterize global performance, the number of yarns remaining valid up to each annotated slice is compared to the number of annotated yarns.
This provides a direct measure of how many trajectories are successfully tracked through the 3D volume. Along with that analysis, we perform a quantitative assessment of each yarn by computing its validity percentage.
This percentage represents the fraction of slices where the yarn’s predicted position lies within the predefined tolerance of the corresponding annotated ground truth.
Only slices for which we dispose of annotations are considered.

\section{Results}
\label{sec:results}
\subsection{Per-slice tracking performance}

\begin{figure}[ht!]
\centering

\begin{subfigure}{0.7\textwidth}
    \centering
    \includegraphics[width=\linewidth]{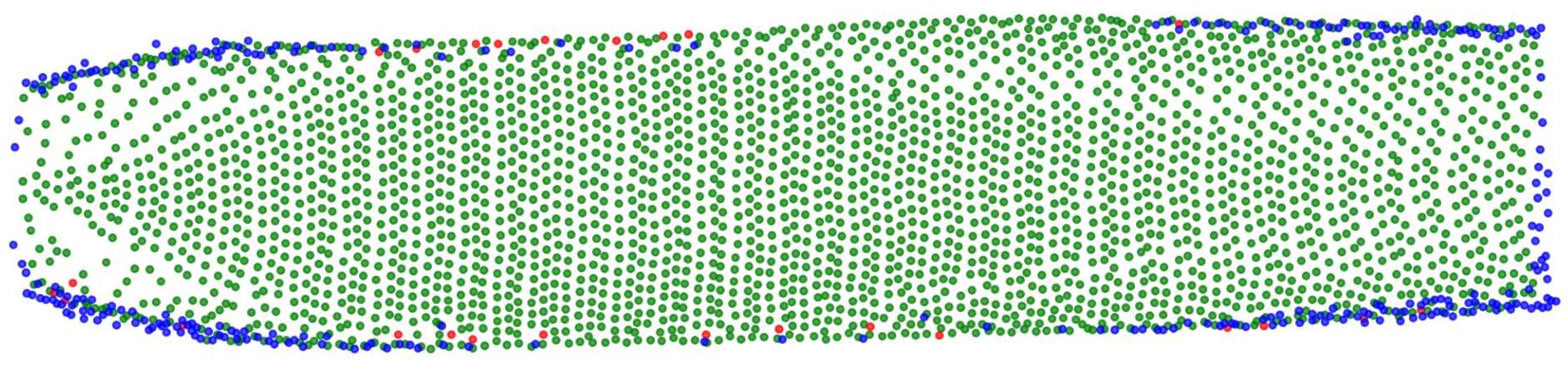}
    \caption{$w=500$}
    \label{fig:result_plan_500_TP}
\end{subfigure}

\vspace{0.7em}

\begin{subfigure}{0.7\textwidth}
    \centering
    \includegraphics[width=\linewidth]{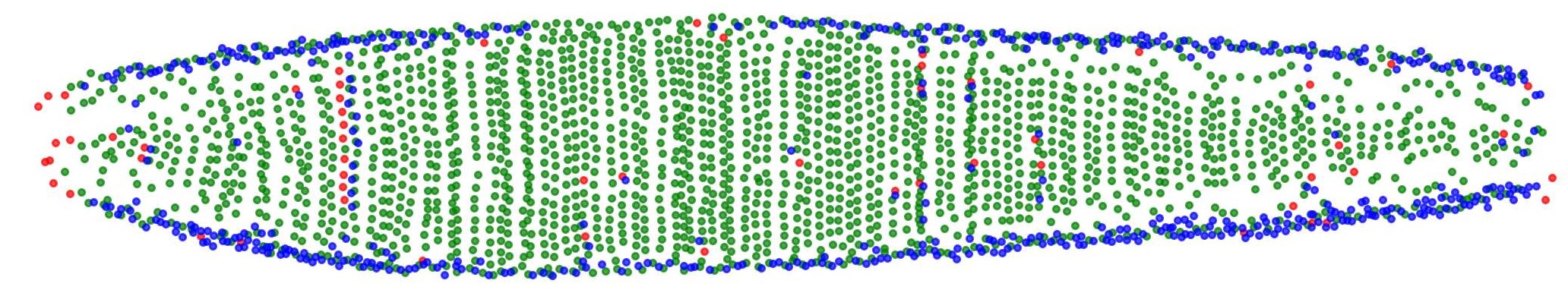}
    \caption{$w=1000$}
    \label{fig:result_plan_1000_TP}
\end{subfigure}

\vspace{0.7em}

\begin{subfigure}{0.7\textwidth}
    \centering
    \includegraphics[width=\linewidth]{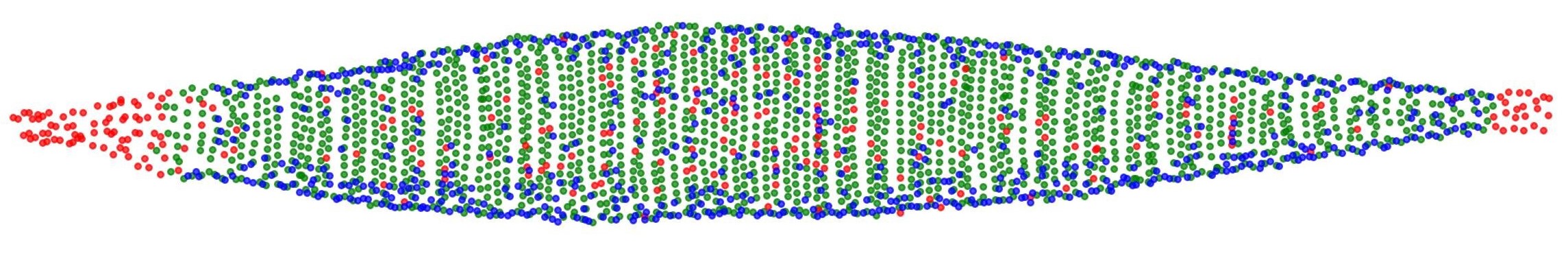}
    \caption{$w=1500$}
    \label{fig:result_plan_00_TP}
\end{subfigure}

\caption{Tracking results on three slices (blue: False Positive, red: False Negative, green: True Positive).}
\label{fig:tracking_results_slices}
\end{figure}

In ~\Cref{tab:yarn_metrics}, we can observe that all performance metrics (sensitivity, F1-score, accuracy and precision) gradually decrease as the distance from the initialization plane increases.
This decline is mainly due to increasing geometric complexity in the airfoil region: yarns that are not present and thus not initialized at the root lead to a high number of false negatives (FN) (\Cref{fig:result_plan_00_TP}), while delayed yarn exits near the part borders (\Cref{fig:result_plan_1000_TP,fig:result_plan_500_TP,fig:result_plan_00_TP}) increase the number of false positives (FP) near the surfaces.

\begin{table}[ht!]
\centering
\caption{Performance metrics of yarn predictions compared to annotations across different slices.}
\label{tab:yarn_metrics}
\small
\begin{tabular}{cccccccc}
\toprule
\textbf{Slice} $w$ & \textbf{TP} & \textbf{FP} & \textbf{FN} & \textbf{Sensitivity} & \textbf{F1-score} & \textbf{Accuracy} & \textbf{Precision} \\
\midrule
500   & 2657 & 284 & 25  & 0.991 & 0.945 & 0.896 & 0.903 \\
1000  & 1954 & 501 & 57  & 0.972 & 0.875 & 0.778 & 0.796 \\
1500  & 1717 & 608 & 253 & 0.872 & 0.800 & 0.666 & 0.738 \\
\bottomrule
\end{tabular}
\end{table}

\subsection{Aggregated tracking performance}

\begin{figure}[ht!]
\centering
\begin{subfigure}[b]{0.48\textwidth}
    \includegraphics[width=\textwidth]{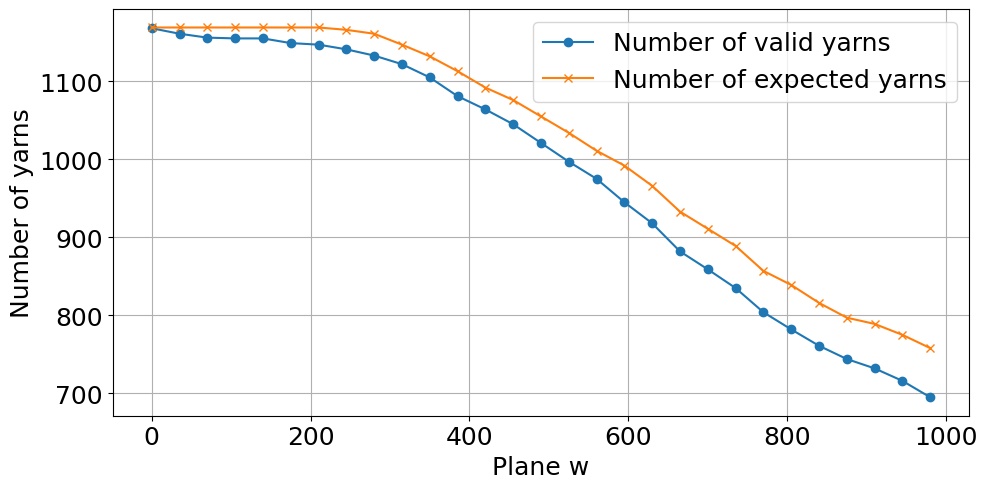}
    \caption{}
    \label{fig:valid_vs_expected}
\end{subfigure}
\hfill
\begin{subfigure}[b]{0.48\textwidth}
    \includegraphics[width=\textwidth]{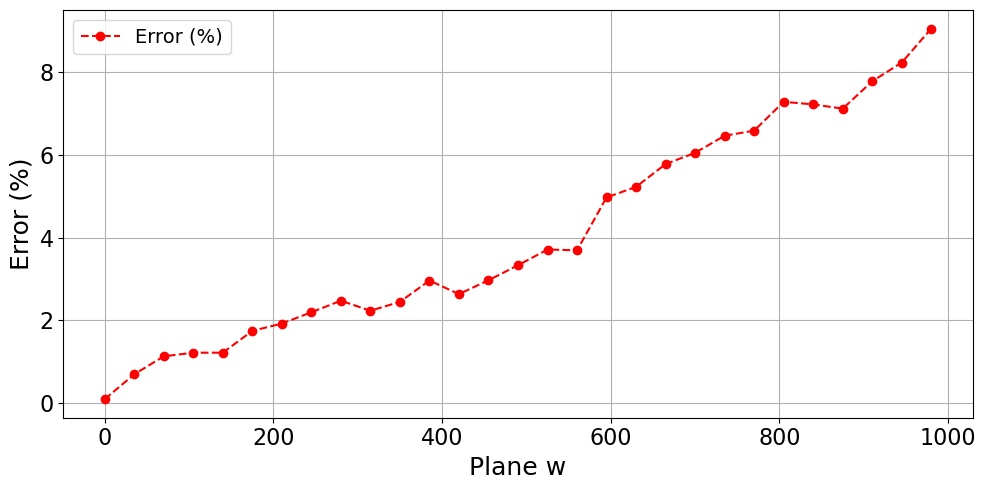}
    \caption{}
    \label{fig:invalid_percent}
\end{subfigure}
\caption{(a) Number of valid yarns vs expected yarns according to their plane $w$, (b) Percentage of invalid yarns according to their plane $w$.}
\label{fig:performance}
\end{figure}

\Cref{fig:performance} presents the evolution of valid versus expected yarns across annotated slices.
As depth increases, the number of valid yarns progressively decreases, reflecting the cumulative nature of tracking errors in our evaluation.
This monotonic increase (\Cref{fig:invalid_percent}) indicates that once an error occurs, it propagates irreversibly along the trajectory. By the final slice, a visible gap appears between the number of expected and valid yarns (\Cref{fig:valid_vs_expected}), corresponding to a cumulative error rate of $8\%$.
This value quantifies the proportion of yarns that lose their trajectory before reaching the end of the volume, providing a global indicator of tracking robustness.

\begin{figure}[ht!]
    \centering
    \includegraphics[width=0.7\textwidth]{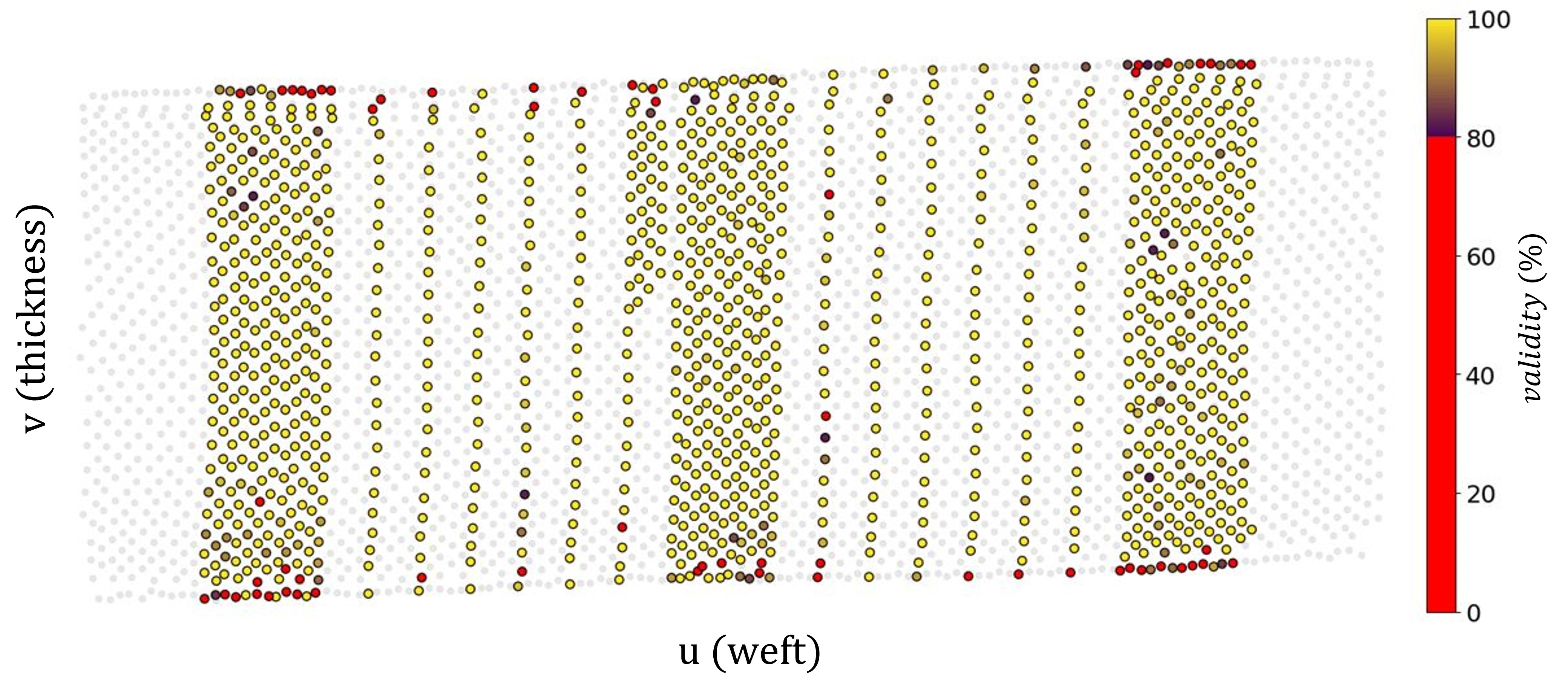}
    \caption{
        Yarns positions at $w=0$.   Light gray points indicate yarns without annotations, while colored points represent annotated yarns, with color corresponding to their validity percentage.
    }
    \label{fig:yarn_validity_slice}
\end{figure}

In~\Cref{fig:yarn_validity_slice}, the positions of the yarns at the slice at $w=0$ are displayed.
Each point is color-coded according to its validity percentage, representing the fraction of slices in which the yarn is correctly matched to the annotated ground truth.
Points with higher percentages are consistently aligned with the ground truth across multiple slices, while lower percentages highlight local mismatches or missing detections.
Notably, yarns with lower validity percentages are located near the boundaries of the part, indicating that tracking errors occur more often in these regions compared to the central areas.

\section{Final Discussion}
\label{sec:conclusion}
\subsection{Conclusion}

This work presented a comprehensive framework for the extraction of all present warp yarn paths from tomographic scans of woven composite preforms.
The proposed method combines local image evidence, geometric regularization, and global structural constraints within a unified variational formulation.
The first two components ensure the accurate localization and smooth tracking of yarn centers across the tomographic volume, while the third component, introduces a novel structural prior that explicitly captures the periodic, crystalline-like organization of the weave.
This integration allows the reconstruction of consistent and physically meaningful yarn trajectories, even in regions where local ambiguities occur.

Applied to a complex fan blade textile reinforcement, the method demonstrated its ability to handle anisotropic yarn geometries, cross-section variability, and interactions between neighboring yarns.
The proposed approach is easily generalizable and can be extended to other woven composite architectures, offering a robust tool for the digital characterization of textile reinforcements directly from volumetric data.

\subsection{Perspectives}

Ongoing efforts to extract the more challenging weft yarns point to the next steps. First, explicit modeling of the warp-weft interactions will prove useful for handling ambiguous or dense regions of yarn crossings. This can be achieved by exploiting the already found yarn paths as repulsion regions to steer away weft yarns within the same probabilistic framework.

Another important element to improve is the static nature of the three properties of interest. Currently, they represent the global (mean) behavior of warp yarns through the entire volume. Nonetheless, these properties are expected to evolve, notably along the warp orientation where the blade  thickness changes the most from root to tip.
In our current algorithm, we only used global (mean) versions of the studied three properties.
In future work, the corresponding energy terms could be adapted to account for spatial variations in yarn behavior within the part.
The architecture evolves significantly from root to tip. Near the root, the weave is denser, yarn cross-sections are more circular, and contacts with weft yarns are more frequent.
Towards the tip, the pattern becomes increasingly sparse, altering local neighborhoods.
Moreover, yarn trajectories adjust depending on their position relative to the upper or lower surfaces of the part, reflecting the influence of the surrounding geometry. Exploiting these location-dependent effects would enable a more realistic modeling of the yarns, thus increasing robustness and resulting in a more stable extraction of the yarn paths.

Finally, the periodic nature of the textile could be exploited, especially in the densest regions.

\section*{Acknowledgments}
Hafsa El Herichi acknowledges the support of a PhD grant N\textsuperscript{o} 2022/1494 from ANRT and Safran. The authors also sincerely thank the anonymous reviewers for their careful reading and constructive comments.

\newpage\appendix
\section{Overall procedure}
\label{sec:procedure_algo}

\begin{algorithm}
\caption{Model parameter estimation}

\textbf{Input} Full 3D tomographic volume $I$, sparse annotations $X'$, $(h,w)$, $(V_r,V_i)$ \\
\textbf{Output} $P_1,P_2,P_3$

\begin{enumerate}

\item \textbf{Estimation of $P_1$}

\begin{itemize}
\item $M \leftarrow$ extract patches $(h,w)$ around each $X'_i$ from $I$
\item $\phi \leftarrow \texttt{PCA}(M)$
\item $P_1 \leftarrow I * \overline{\phi}$ (Eq. 1)
\end{itemize}

\item \textbf{Estimation of $P_2$}

\begin{itemize}
\item $P_2 \leftarrow \texttt{estimation\_gmm}(X',1)$ (Eq. 2)
\end{itemize}

\item \textbf{Estimation of $P_3$}

\begin{itemize}
\item $X_r \leftarrow \texttt{subset\_neighbors}(X', V_r)$
\item $P_3 \leftarrow \texttt{estimation\_gmm}(X_r, V_r)$ (Eq.~3)
\end{itemize}
\end{enumerate}

\end{algorithm}

\begin{algorithm}
\caption{Tracking on one slice: \texttt{tracking\_one\_slice}}

\textbf{Input} Positions at previous slice $X(w-1)$, current tomographic slice $I_w$, parameters of defined statistical properties $(P_1,P_2,P_3)$, $(\alpha,\beta)$ \\
\textbf{Parameters} $max\_iter=5$, $conv\_crit=10^{-6}$ \\
\textbf{Output} $X(w)$

\begin{enumerate}

\item $X(w) \leftarrow X(w-1)$

\item Repeat while $iter < max\_iter$ and $\|dX\| > conv\_crit$

\begin{itemize}

\item $\nabla E_1,\nabla E_2,\nabla E_3 \leftarrow$ (Eq. 12), (Eq. 13), (Eq. 14)

\item $\nabla E \leftarrow \nabla E_1 + \alpha \nabla E_2 + \beta \nabla E_3$ (Eq. 5)

\item $H_1,H_2,H_3 \leftarrow$ (Eq. 15), (Eq. 16), (Eq. 17)

\item $K \leftarrow H_1 + \alpha H_2 + \beta H_3$ (Eq. 6)

\item $dX \leftarrow -K^{-1}\nabla E$

\item $X(w) \leftarrow X(w) + dX$

\item $X(w) \leftarrow \texttt{fix\_positions}(X(w))$

\end{itemize}

\end{enumerate}

\end{algorithm}

\begin{algorithm}
\caption{Volume tracking}

\textbf{Input} Full 3D tomographic volume $I$ \\
\textbf{Output} Full computed yarn paths $X$

\begin{enumerate}

\item \textbf{Initialization}

\begin{itemize}
\item $X(w_0) \leftarrow \texttt{local\_maxima}(I_{w_0} * \overline{\phi})$
\item $X(w_0) \leftarrow \texttt{fix\_positions}(X(w_0))$
\item $X(w_0) \leftarrow \texttt{manual\_corrections}(X(w_0))$
\end{itemize}

\item \textbf{Slice-by-slice tracking}

For each slice $w = 1,\dots,w_{end}$:

\begin{itemize}
\item $X(w) \leftarrow \texttt{tracking\_one\_slice}(X(w-1),I_w)$
\end{itemize}

\end{enumerate}

\end{algorithm}

\section{Annotation Study}
\label{sec:appendix_study}

To estimate the required number of annotations, a densely annotated sub-volume (250 yarns, 30 slices) is used as reference to simulate reduced scenarios with fewer yarns or slices. The statistics $P_1, P_2,$ and $P_3$ are compared using the $L_2$ norm.
Reliable estimates are obtained with $\approx$100 yarns and 20–-25 slices. Histogram variations affect the resulting GMMs and require adjusting the variational weighting.

\begin{figure}[ht]
\centering

\begin{subfigure}{0.48\linewidth}
\centering
\includegraphics[width=1\linewidth]{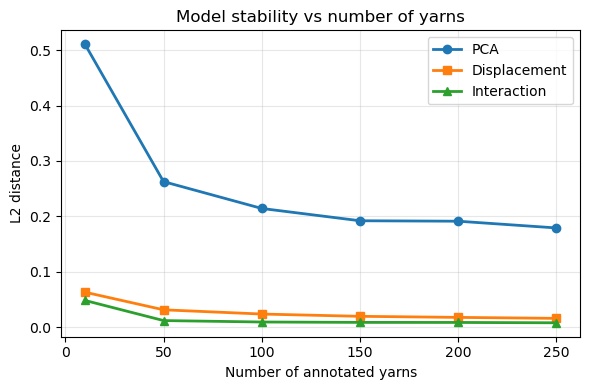}
\caption{Protocol 1: varying the number of annotated yarns (30 slices).}
\label{fig:protocol1_global}
\end{subfigure}
\hfill
\begin{subfigure}{0.48\linewidth}
\centering
\includegraphics[width=1\linewidth]{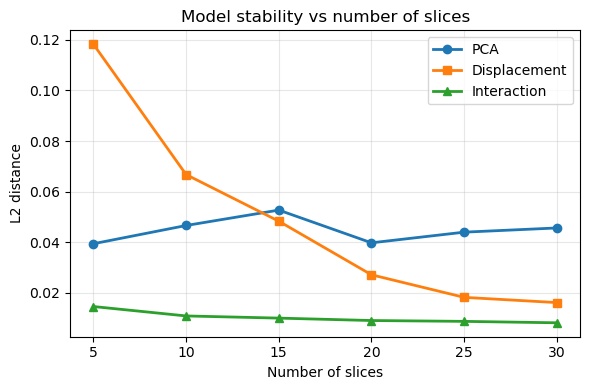}
\caption{Protocol 2: varying the number of annotated slices (250 yarns).}
\label{fig:protocol2_global}
\end{subfigure}

\caption{L$_2$ error of the three statistical analyses for both annotation reduction protocols, relative to the full-annotation reference.}

\label{fig:global_protocols}
\end{figure}

\end{document}